\documentclass[fleqn,10pt]{wlscirep}
\usepackage[utf8]{inputenc}
\usepackage[T1]{fontenc}
\title{Filtering out mislabeled training instances using black-box optimization and quantum annealing}
\usepackage{subfigure}
\usepackage{algorithm}
\usepackage{algorithmicx}
\usepackage{algpseudocode}
\usepackage{bm}
\usepackage{soul}

\author[1,2*]{Makoto Otsuka}
\author[2]{Kento Kodama}
\author[2,3]{Keisuke Morita}
\author[2,3,4,5]{Masayuki Ohzeki}
\affil[1]{LiLz Inc., Okinawa, Japan}
\affil[2]{Graduate School of Information Sciences, Tohoku University, Miyagi, Japan}
\affil[3]{Sigma-i Co., Ltd., Tokyo, Japan}
\affil[4]{Department of Physics, Institute of Science Tokyo, Tokyo, Japan}
\affil[5]{Research and Education Institute for Semiconductors and Informatics, Kumamoto University, Kumamoto, Japan}

\affil[*]{m.otsuka@lilz.jp}


\begin{abstract}
This study proposes an approach for removing mislabeled instances from contaminated training datasets by combining surrogate model-based black-box optimization (BBO) with postprocessing and quantum annealing.
Mislabeled training instances, a common issue in real-world datasets, often degrade model generalization, necessitating robust and efficient noise-removal strategies.
The proposed method evaluates filtered training subsets based on validation loss, iteratively refines loss estimates through surrogate model-based BBO with postprocessing, and leverages quantum annealing to efficiently sample diverse training subsets with low validation error.
Experiments on a noisy majority bit task demonstrate the method's ability to prioritize the removal of high-risk mislabeled instances.
Integrating D-Wave's clique sampler running on a physical quantum annealer achieves faster optimization and higher-quality training subsets compared to OpenJij's simulated quantum annealing sampler or Neal's simulated annealing sampler, offering a scalable framework for enhancing dataset quality.
This work highlights the effectiveness of the proposed method for supervised learning tasks, with future directions including its application to unsupervised learning, real-world datasets, and large-scale implementations.
\end{abstract}
\begin{document}

\flushbottom
\maketitle
\thispagestyle{empty}

\section*{Introduction}

Machine learning is a general framework for extracting underlying patterns contained in data and making accurate predictions or reasonable decisions in new but similar situations based on the learned patterns~\cite{bishop2006pattern, murphy2012machine}.
Although many machine learning models have been proposed for various problem settings, most assume that the given data are correctly annotated and that the given target labels are noise-free.
However, the given dataset is often noisy in real-world applications and contains mislabeled instances due to human errors or other unforeseen reasons~\cite{lee2021survey, chu2016data}.
These mislabeled instances can significantly degrade the generalization performance of the trained model, highlighting the critical need for robust techniques to handle noisy data~\cite{brodley1999identifying,smith2011improving, smith2014potential}.

Numerous studies have demonstrated that removing noisy or harmful training instances can significantly improve model performance on contaminated datasets.
For example, removing outliers or mislabeled instances before training has improved classification accuracy~\cite{brodley1999identifying, smith2011improving}.
Moreover, the benefits of eliminating harmful training instances often surpass those of hyperparameter optimization~\cite{smith2014potential}.
Traditional techniques have relied on outlier detection methods, noise removal techniques, or ensembles of different algorithms to eliminate mislabeled instances~\cite{smith2013extensive}.
Another approach involves removing instances that are mostly misclassified by a set of classifiers~\cite{brodley1999identifying, smith2014potential}.
Furthermore, some studies indicate that noise in class labels has a greater negative impact on the performance of the induced model than noise in input attributes~\cite{zhu2004class, nettleton2010study}.
All of these studies indicate that the removal of detrimental instances significantly improves the quality of the learned model.

In recent years, more theoretically grounded techniques have been proposed to address label noise in large-scale datasets~\cite{song2022learning, li2024noisy}.
For instance, influence functions~\cite{koh2017understanding} estimate how much individual samples affect model predictions, enabling principled identification of harmful points.
Co-teaching strategies~\cite{han2018co} train two networks simultaneously, each filtering out potentially noisy instances based on peer loss.
Other approaches rely on training dynamics such as forgetting events~\cite{toneva2018empirical}, confidence variability~\cite{swayamdipta2020dataset}, or margin tracking~\cite{pleiss2020identifying} to identify mislabeled data.
These approaches are particularly well-suited for large-scale datasets, as they exploit patterns in model behavior that are indicative of label noise, making them practical and effective in real-world settings.

However, most of these existing approaches do not directly optimize validation loss.
This is despite the fact that validation loss serves as a practical and accessible proxy for the generalization loss, which is the ultimate goal in supervised learning.
If the validation loss could be directly optimized, the training subset would be refined in a more comprehensive way.
It would not only filter out mislabeled instances but also address class imbalance, eliminate samples with unreliable features, and mitigate other factors detrimental to generalization.
Direct optimization of validation loss is theoretically principled and practically appealing.
However, this approach is seldom explored in the field because identifying the optimal training subset with respect to validation loss amounts to solving an inherently difficult combinatorial optimization problem.

In recent years, surrogate model-based black-box optimization (BBO)~\cite{baptista2018bayesian,koshikawa2021benchmark,koshikawa2021combinatorial,doi2023exploration} has gained attention as an efficient strategy for searching combinatorial spaces where the objective function is not explicitly known.
The surrogate model constructs a quadratic approximation of the unknown objective function based on previously evaluated points.
This approximation enables efficient selection of the next candidate solution in the combinatorial space.
In particular, surrogate model-based BBO has been shown to work effectively in combination with annealing-based samplers~\cite{baptista2018bayesian}, especially well with quantum annealing samplers~\cite{koshikawa2021benchmark,koshikawa2021combinatorial,doi2023exploration} running on physical quantum annealers like D-Wave Advantage~\cite{mcgeoch2020}.
Quantum annealing~\cite{kadowaki1998, farhi2001quantum, johnson2011quantum} has traditionally been applied to solve combinatorial optimization problems~\cite{mcgeoch2013experimental, lucas2014ising}, but it is also well suited for identifying high-quality candidate points due to its demonstrated capability to sample diverse and near-optimal solutions efficiently~\cite{zucca2021diversity, mohseni2023sampling}.
However, this method has not yet been explored in the context of the dataset cleansing or noisy-label removal.
Although the number of qubits available on the current quantum annealers remains limited to the order of thousands as of 2025 and not ready to handle real-world datasets yet, this number is steadily increasing.
This study aims to provide preliminary insights into the future applicability of quantum-assisted optimization for data quality improvement, using a well-controlled experimental setup.

In this work, we apply the surrogate model-based BBO with postprocessing~\cite{morita2023random} for cleansing contaminated training datasets, and evaluated its effectiveness using a well-controlled experiment called a noisy majority bit task.
In the postprocessing phase, previously explored solutions are discarded from the set of sampled candidates. If a high-quality, unexplored solution is available, it is selected as the next candidate. Otherwise, a random sample is chosen to encourage further exploration.
The performance of our approach is assessed using three different annealing based-samplers: the simulated annealing sampler, the simulated quantum annealing sampler, and quantum annealing sampler operating on D-Wave's actual quantum hardware.
We hypothesize that the diversity of samples generated by the quantum annealer yields a broader set of near-optimal solutions, including those not previously explored, thereby outperforming classical simulators in this context.
Our approach utilizes (1) a validation loss metric for selecting data points under the assumption of a clean validation dataset, (2) a sequential training subset selection strategy utilizing BBO with a surrogate model estimating validation loss, and (3) the unique capabilities of quantum annealers to sample diverse and high-quality solutions efficiently. Together, these components provide a principled and effective framework for addressing the challenge of noise-label removal.

The remainder of this paper is organized as follows.
In the next section, we first describe the noisy majority bit task designed to evaluate the proposed method and then explain the proposed method in detail with its two building blocks: the surrogate model-based BBO with postprocessing and energy-based samplers such as D-Wave's quantum annealer.
In the following section, we show experimental results with unique characteristics of the proposed method, including prioritized removal of high-risk mislabeled instances, iterative improvement of solution quality, and its ability to leverage D-Wave's quantum annealer to reach better solutions faster.
In the final section, we discuss the insights gained from this study and outline potential directions for future work.

\section*{Methods}

\subsection*{Noisy majority bit task}
To evaluate the effectiveness of the proposed algorithm in removing mislabeled instances, we designed a classification task called the noisy majority bit task.
This task aims to predict the majority bit in a given binary input vector.
The dataset for this task is partitioned into distinct, non-overlapping training, validation, and test sets to ensure an unbiased evaluation.
For instance, the correct target label is $0$ for the binary input vector $[0, 0, 0, 1, 0, 1, 0, 0, 1]$, whereas $1$ for $[1, 1, 0, 1, 0, 1, 1, 0, 0]$.
Under ideal conditions, where all training instances are accurately labeled according to the majority bit, this task would be straightforward.
However, the actual training set includes noisy instances where the labels are flipped to the minority bit, introducing mislabeled data.
This contamination makes the noisy majority bit task non-trivial, providing a controlled and challenging framework for evaluating the proposed algorithm's effectiveness in noise removal.

\begin{figure}
\centering
\includegraphics[width=\linewidth]{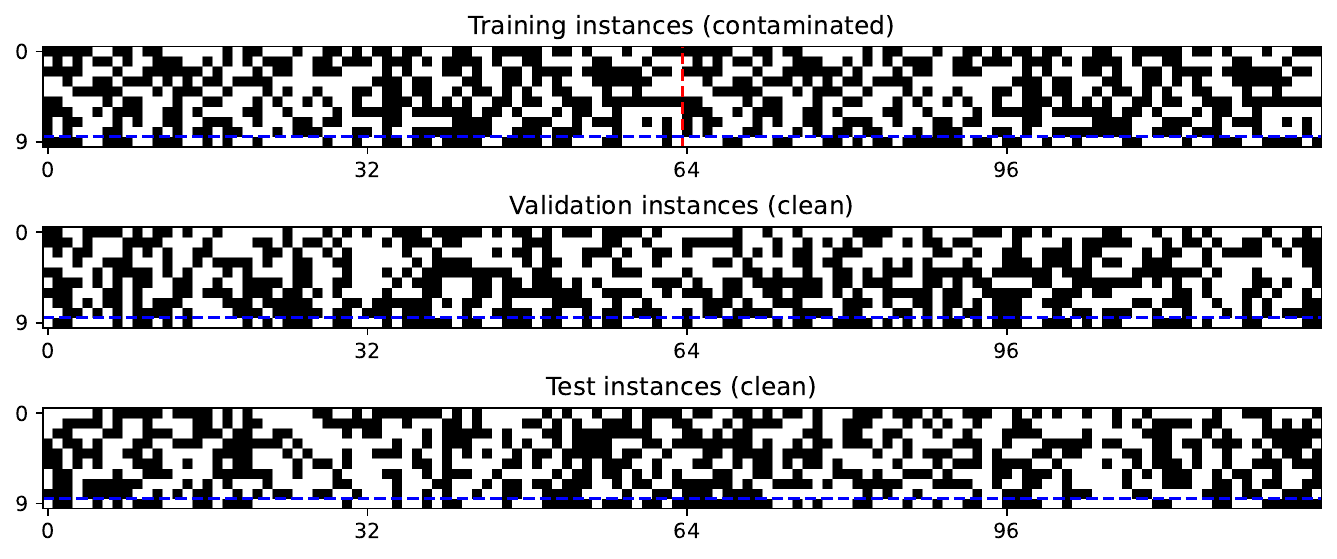}
\caption{
Training, validation and test datasets for the noisy majority bit task.
For any dataset, each column represents a specific instance composed of 9 input bits $\mathbf{x} \in \{0, 1\}^{9}$ and 1 target bit $t \in \{0, 1\}$, which are separated by the horizontal dashed blue line.
64, 128, and 128 non-overlapping binary patterns were randomly sampled without repetition to create the input patterns for the training, validation, and test sets, respectively.
Each input pattern in the validation and test sets was correctly labeled based on the majority bit of the 9 input bits.
The training dataset consists of two halves, separated by a vertical dashed red line.
The first half contains 64 real training instances correctly labeled by the majority bits, while the latter half contains 64 fake training instances incorrectly labeled by the minority bits.
These incorrectly-labeled fake training instances should be removed for successful training.
}
\label{fig:contaminated_train_valid_test_instalces}
\end{figure}

The training, validation, and test sets were constructed as follows.
Nine-dimensional input feature vectors for these sets were generated by randomly sampling 64, 128, and 128 unique binary vectors, respectively.
All instances in the validation and test sets were correctly labeled according to their majority bits, ensuring clean datasets for evaluation.
For the training set, the 64 input vectors were used to create 128 labeled training instances: 64 correctly labeled (\textit{real}) instances with majority-bit labels and 64 mislabeled (\textit{fake}) instances with minority-bit labels.
An overview of these datasets, used in all experiments, is shown in Fig.~\ref{fig:contaminated_train_valid_test_instalces}.

In this task, the assumption of a clean validation set is critical, as detrimental instances in the training set cannot be identified or removed without the additional information provided by the validation set.
Without filtering out the mislabeled training instances, no base model can achieve better-than-chance performance based solely on the contaminated training set.
Therefore, the successful removal of mislabeled instances is essential for training a base model that performs effectively on the test set.

By reducing the validation loss against the noise-free validation set, we can remove potentially detrimental training instances from the training dataset.
By assuming the correctly-labeled validation dataset, we can use the validation loss as a proxy for the generalization performance of the trained model.
In other words, we can evaluate how good the particular training subset is by the validation loss of the model trained on the filtered training subset.

\subsection*{Surrogate model-based BBO with postprocessing}
Although validation loss can be used as a proxy for the generalization performance of a trained model, it is computationally prohibitive to evaluate the validation loss for all possible subsets of the original training dataset.
To address this issue, we employed surrogate model-based BBO~\cite{baptista2018bayesian,koshikawa2021benchmark,koshikawa2021combinatorial,doi2023exploration} with postprocessing~\cite{morita2023random} to efficiently identify training subsets with low validation error in an iterative manner.
Although surrogate model-based BBO with postprocessing is not itself novel, its application to noise removal---combined with the use of diverse samples from a quantum annealer during the acceptance process---is a key contribution of this work.
The following part explains the detailed procedure of this algorithm and its application to the removal of mislabeled training instances.
The pseudo code of the algorithm is shown in Algorithm \ref{alg:bbo}.

\begin{algorithm}
\caption{Mislabeled instance removal algorithm using surrogate model-based BBO with postprocessing}\label{alg:bbo}
\begin{algorithmic}[1]
\renewcommand{\algorithmicrequire}{\textbf{Input:}}
\renewcommand{\algorithmicensure}{\textbf{Output:}}
\Require $D_{\mathrm{train}} \equiv \{(\bm{x}_{1}, t_{1}), \ldots, (\bm{x}_{n}, t_{n}) \}, D_{\mathrm{valid}}$, QUBO-based sampler, Number of samples drawn from the sampler $M$, Total number of steps $N$; Number of initialization steps: $N_{0}$, Regression model $R$, Supervised-learning model $P$, Monotonically increasing function $g$
\Ensure $\bm{q}_{k^{*}}$ with the lowest validation loss
\For{$k = 1$ to $N$}
\If{$k \leq N_{0}$}
\State Randomly sample $\bm{q}_{k} \in \{0, 1\}^{n}$
\Else
\State Compute coefficient $\tilde{\bm{\alpha}}_{k} \in \mathbb{R}^{p}$ with the regression model: $\tilde{\bm{\alpha}}_{k} = R(\bm{\tilde{q}}_{1:k-1}, \bm{l}_{1:k-1})$
\State Sample $\{\bm{q}_{k}^{(m)}\}_{m=1}^{M}$ from the sampler using the QUBO $\bm{U}(\tilde{\bm{\alpha}}_{k}) \in \mathbb{R}^{n \times n}$
    \If{all samples in $\{\bm{q}_{k}^{(m)}\}_{m=1}^{M}$ is already contained in the evaluated set $S_{k-1} \equiv \{\bm{q}_{1}, \ldots, \bm{q}_{k-1}\}$}
        \State Randomly sample $\bm{q}_{k} \in \{0, 1\}^{n}$ not included in the evaluated set $S_{k-1}$
    \Else
        \State Select the sample with the lowest energy in $\{\bm{q}_{k}^{(m)}\}_{m=1}^{M}$ as $\bm{q}_{k} \in \{0, 1\}^{n}$
    \EndIf
\EndIf
\State Create a filtered training subset 
$D_{\mathrm{train}}(\bm{q}_{k})$
\State Train a supervised learning model $P$ with $D_{\mathrm{train}}(\bm{q}_{k})$ to get the trained model $P^{*}$
\State Evaluate the trained model $P^{*}$ with $D_{\mathrm{valid}}$ to get the validation loss $L_{\mathrm{valid}}(\bm{q}_{k})$
\State Compute $l_{k} = g(L_{\mathrm{valid}}(\bm{q}_{k}))$ where $g$ is a monotonically increasing function
\State Keep $\bm{q}_{k} \in \{0, 1\}^{n}$ as an evaluated set $S_{k} \equiv \{\bm{q}_{1}, \ldots, \bm{q}_{k}\}$
\State Keep $\tilde{\bm{q}}_{k} \in \{0, 1\}^{p}$ as a matrix $\bm{\tilde{q}}_{1:k} \in \{0, 1\}^{k \times p}$
\State Keep $l_{k} \in \mathbb{R}$ as a vector $\bm{l}_{1:k} \in \mathbb{R}^{k}$
\EndFor
\end{algorithmic}
\end{algorithm}

First, let us denote the training, validation, and test sets as $D_{\mathrm{train}}$, $D_{\mathrm{valid}}$, and $D_{\mathrm{test}}$, respectively.
The potentially contaminated original training dataset is defined as $D_{\mathrm{train}} \equiv \{(\bm{x}_{i}, t_{i})\}_{i=1}^{n} \equiv \{(\bm{x}_{1}, t_{1}), \ldots, (\bm{x}_{n}, t_{n})\}$
where the $i$-th tuple $(\bm{x}_{i}, t_{i})$ is called the $i$-th training instance, which is defined as a pair of $i$-th input feature vector $\bm{x}_{i}$ and $i$-th target label $t_{i}$.
Target labels are assumed to be noisy in $D_{\mathrm{train}}$
but noise-free in both $D_{\mathrm{valid}}$ and $D_{\mathrm{test}}$.

Second, let us define an $n$-dimensional binary selection vector $\bm{q} \in \{0, 1\}^{n}$ to denote a filtered training subset $D_{\mathrm{train}}(\bm{q}) \subseteq D_{\mathrm{train}}$.
The $i$-th training instance $(\bm{x}_{i}, t_{i}) \in D_{\mathrm{train}}$ is included in a filtered training subset $D_{\mathrm{train}}(\bm{q}) \subseteq D_{\mathrm{train}}$ if $q_{i} = 1$ and excluded if $q_{i} = 0$.
Thus, a binary pattern $\bm{q}$ uniquely determines a specific filtered subset $D_{\mathrm{train}}(\bm{q})$ from the original training dataset $D_{\mathrm{train}}$.
Consequently, there are $2^{n}$ possible realizations of filtered training subsets $D_{\mathrm{train}}(\bm{q})$ in total.

Third, we introduce the surrogate model $f_{\tilde{\bm{\alpha}}}(\bm{q})$, which is defined as
\begin{align}
f_{\tilde{\bm{\alpha}}}(\bm{q}) 
=&
\alpha_{0} + 
\bm{q}^{\top} \bm{U}(\tilde{\bm{\alpha}}_{\backslash 0}) \bm{q}
\label{eq:surrogate_1}
\\
=&
\alpha_{0} + 
\bm{q}^{\top} 
\begin{bmatrix}
\alpha_{1} & \alpha_{1, 2} & \cdots & \alpha_{1, n} \\ 
0 & \alpha_{2} & \cdots & \alpha_{2, n} \\ 
\vdots & \vdots & \ddots & \vdots \\ 
0 & 0 & \cdots & \alpha_{n} \\ 
\end{bmatrix}
\bm{q}
\label{eq:surrogate_2}
\\
=&
\alpha_{0} + \sum_{j} \alpha_{j} q_{j} + \sum_{i < j} \alpha_{i,j} q_{i} q_{j}
\label{eq:surrogate_3}
\\
=& 
[\alpha_{0}, \alpha_{1}, \ldots, \alpha_{n}, \alpha_{1, 2}, \alpha_{1, 3}, \ldots, \alpha_{n-1, n}] 
[1, q_{1}, \ldots, q_{n}, q_{1} q_{2},  q_{1} q_{3}, \ldots, q_{n-1} q_{n}]^{\top}
\label{eq:surrogate_5}
\\
=& \tilde{\bm{\alpha}}^{\top} \tilde{\bm{q}}
\label{eq:surrogate_6}
\end{align}
to estimate the validation loss of the model trained by the filtered training subset $D_{\mathrm{train}}(\bm{q})$.
In the surrogate model-based BBO, the surrogate model plays two roles: one is to approximate the unknown cost function with a quadratic binary form and the other is to generate binary samples with sufficiently low cost values.
Here, the surrogate model $f_{\tilde{\bm{\alpha}}}(\bm{q})$ is parameterized by the parameter vector $\tilde{\bm{\alpha}} \in \mathbb{R}^{p}$ given in the form of Eq.~\eqref{eq:surrogate_5} where $p = 1 + n + n (n - 1) / 2$.
The elements in this parameter vector $\tilde{\bm{\alpha}}$ are also present in the constant-shifted quadratic unconstrained binary optimization (QUBO) form in Eq.~\eqref{eq:surrogate_2}.
Eq.~\eqref{eq:surrogate_2} is composed of the first term $\alpha_{0}$ which represents a constant shift and the second quadratic term which represents linear interaction by its diagonal elements and pair-wise interaction by its off-diagonal elements.
In short, the unknown true cost function $f(\bm{q})$ over $\bm{q} \in \{0, 1\}^{n}$ is approximated by the second order Taylor expansion around $\bm{q}$, which is $f_{\tilde{\bm{\alpha}}}(\bm{q})$ in the form shown in Eq.~\eqref{eq:surrogate_3}.
The quality of this approximation can be improved by adjusting the parameter vector $\tilde{\bm{\alpha}}$.

Fourth, we sample low-cost binary patterns $\bm{q}$ from the constant-shifted surrogate model $f_{\tilde{\bm{\alpha}}}(\bm{q}) - \alpha_{0}$, which takes the quadratic form as shown in Eq.~\eqref{eq:surrogate_1} and we can use any QUBO-based sampler such as a sampler running on D-Wave System's quantum annealer.
The sampler generates a binary pattern $\bm{q}$ with sufficiently low cost value $f_{\tilde{\bm{\alpha}}}(\bm{q})$ under the fixed parameter vector $\tilde{\bm{\alpha}}$.
$M$ samples are drawn simultaneously from a sampler in a single sampling step.

Fifth, we update the parameter vector $\tilde{\bm{\alpha}}$ to minimize the validation loss of the model trained by the filtered training subset $D_{\mathrm{train}}(\bm{q})$.
This optimization can be done in two phases.
The first phase updates the parameter vector $\tilde{\bm{\alpha}}$ using the random samples in a batch manner.
The second phase updates the parameter vector $\tilde{\bm{\alpha}}$ in an online manner using all samples accepted up to the current step.

In the first phase, we randomly sample the $n$-dimensional binary pattern $\bm{q}_{k} \in \{0, 1\}^{n}$ sequentially from $k = 1$ to $k = N_{0}$,
and then for each binary pattern $\bm{q}_{k}$, we compute the validation loss $L_{\mathrm{valid}}(\bm{q}_{k})$ of the model trained by the filtered training subsets $D_{\mathrm{train}}(\bm{q}_{k})$.
The validation loss $L_{\mathrm{valid}}(\bm{q}_{k})$ might not be distributed normally, and it is often preferable to transform it by a monotonically increasing function $g$ such as the log function to make it more normally distributed when this value becomes the target of some model and this is the case in our algorithm.
Let us define the transformed validation loss as $l_{k} = g(L_{\mathrm{valid}}(\bm{q}_{k}))$.
If we define $\tilde{\bm{q}}$ as in Eq.~\eqref{eq:surrogate_5}, the transformed validation loss $l_{k}$ can be approximated by the surrogate model $f_{\tilde{\bm{\alpha}}}(\bm{q}_{k})$ in the form of Eq.~\eqref{eq:surrogate_6}, which is linear in $\tilde{\bm{q}}$.
Therefore, if we collect the transformed validation losses $\{l_{1}, \ldots, l_{N_{0}}\}$ and the corresponding binary patterns $\{\bm{q}_{1}, \ldots, \bm{q}_{N_{0}}\}$, we can update the parameter vector $\tilde{\bm{\alpha}}$ to minimize the transformed validation losses $\{l_{1}, \ldots, l_{N_{0}}\}$.
This optimization can be done in many forms such as the Bayesian linear regression with sparse prior~\cite{baptista2018bayesian,carvalho2009handling,bhattacharya2022geometric}, the LASSO regression~\cite{tibshirani1996regression}, and the Ridge regression to update the parameter vector $\tilde{\bm{\alpha}}$ depending on the sparsity assumption of the given problem.
This first step is necessary to obtain a good initial guess of the parameter vector $\tilde{\bm{\alpha}}$.

In the second phase, we sequentially accept one good sample per step
and compute the parameter vector $\tilde{\bm{\alpha}}_{k}$ with sequentially updated input-output pairs $\{(\tilde{\bm{q}}_{j}, l_{j})\}_{j=1}^{k-1}$ available at each time step $k \in \{N_{0} + 1, N_{0} + 2, \ldots, N\}$.
The good sample is selected as follows.
After drawing $M$ samples $\{\bm{q}_{k}^{(m)}\}_{m=1}^{M}$ from a sampler at step $k$, the one with the lowest energy that is not included in the accepted sample set $S_{k-1} \equiv \{\bm{q}_{1}, \ldots, \bm{q}_{k-1}\}$ is selected.
If no such sample exists, a random sampling is conducted until a previously unseen sample is found, which is then accepted.
The accepted sample $\bm{q}_{k}$ at step $k$ is then used to compute the transformed validation loss $l_{k}$ at step $k$, by transforming the validation loss $L_{\mathrm{valid}}(\bm{q}_{k})$ of the model trained by a filtered training subset $D_{\mathrm{train}}(\bm{q}_{k})$.
Newly obtained quantities at time step $k$ such as $\bm{q}_{k} \in \{0, 1\}^{n}$, $\tilde{\bm{q}}_{k} \in \{0, 1\}^{p}$, $l_{k} \in \mathbb{R}$ are stored for the next iteration.
At the end of the second phase, the best sample $q_{k^{*}}$ with the lowest transformed validation cost is reported where $k^{*} = \mathrm{argmin}_{k} l_{k}$.

In our experiment, we used the following parameters and settings.
$n=128$, $N_{0} = 64$, $N = 320 (= 64 + 256)$, $M = 512$, $p = 8257 = 1 + 128 + 128 \times (128 - 1) / 2$.
We used a logistic regression model as a task-specific base model for solving the noisy majority bit task.
The reason for using this model is that it is simple, interpretable, and fast to train.
As a linear model for estimating $\tilde{\bm{\alpha}}$, we used the ridge regression model instead of the Bayesian model with sparsity prior or the Lasso model, which is usually used in the context of the surrogate model-based BBO to impose the sparsity constraint on the QUBO matrix.
The reason for not using the sparse version of the model is that the data is generated in such a way that half of the training data points have noisy labels, and we expect the resulting QUBO matrix can be well approximated without sparsity assumption.

\subsection*{Complexity analysis}

The computational cost of the proposed algorithm mainly arises from training the surrogate model and executing the sampling loop.
The surrogate model involves $p = O(n^{2})$ parameters for a training dataset of size $n$.
At each optimization step $k$ with $N_{0} < k \le N$, the regression update uses all $k-1$ previously evaluated samples, which requires operations on a $(k \times p)$ data matrix.
In the worst case, solving ridge regression from scratch at step $k$ costs $O(kp^{2} + p^{3})$.
Summed over $N$ optimization steps, this yields a total worst-case complexity of $O(N^{2} p^{2} + N p^{3})$, or equivalently $O(N^{2} n^{4} + N n^{6})$.

The sampling loop contributes an additional $O(N M S n)$ operations, where $M$ is the number of reads (samples) drawn per step and $S$ is the number of sweeps per read.
For classical samplers (Neal SA or OpenJij SQA), one read with a fixed $S$ visits all $n$ binary variables per sweep with incremental energy updates, yielding a per-read cost of $O(S n)$ and a total sampling cost of $O(N M S n)$. 
In contrast, for the D-Wave QA clique sampler, the per-step wall-clock time (including embedding) is nearly constant, so its sampling cost scales as $O(N)$.
These asymptotic bounds are consistent with the measured mean per-step times in Fig.~\ref{fig:sampling_time_statistics} ($1.196$\,s for OpenJij SQA, $0.104$\,s for Neal SA, and $0.011$\,s for D-Wave QA clique), which show that sampling dominates the runtime for classical samplers under our settings, whereas the quantum sampler contributes only a small sampling overhead.

Although substituting $(n, N, M, S)=(128, 320, 512, 1000)$ into the asymptotic forms may suggest that the surrogate term $O(N^{2} n^{4} + N n^{6})$ dominates $O(N M S n)$, wall-clock runtime is governed by constant factors and implementation efficiency.
In our setting, ridge regression is performed using the highly optimized \texttt{scikit-learn} implementation, which results in small constant factors. 
By contrast, SA/SQA must generate $M$ reads with $S$ sweeps per read, incurring substantial per-read overhead.
Consequently, the sampling loop is the primary bottleneck for classical samplers, as shown in Fig.~\ref{fig:sampling_time_statistics}.

\subsection*{D-Wave's quantum annealer and related samplers}

Due to the combinatorial nature of the problem, it is difficult to find the optimal solution in a reasonable amount of time.
To address this issue, we propose to leverage the characteristics of D-Wave Systems' quantum annealer, which is known to provide near-optimal and diverse samples quickly~\cite{zucca2021diversity, mohseni2023sampling}.

We used three samplers in this study.
The first sampler is the OpenJij's simulated quantum annealing (SQA) sampler, which is a classical sampler provided by the private company called Jij~\cite{openjij}.
The second sampler is the Neal's simulated annealing (SA) sampler, another classical sampler offered by D-Wave Systems~\cite{dwaveneal}.
We selected Neal's SA sampler over OpenJij because of their comparable performance characteristics\cite{comparison2025}, thereby ensuring a fair comparison with the subsequent sampler also offered by D-Wave Systems.
The third sampler is the D-Wave's clique sampler~\cite{dwavecliquesampler}, running on the physical quantum annealer called the D-Wave Advantage 6.4~\cite{mcgeoch2020}.
In the following section, we will refer to these samplers as the OpenJij SQA sampler, the Neal SA sampler, and D-Wave QA clique sampler, respectively.

For all three samplers, we used the latest version of the respective Python packages available at the time and employed their default parameters for class instantiation without modifications unless otherwise specified.
For the OpenJij SQA sampler, we used the \texttt{openjij.SQASampler} class from the \texttt{openjij} package (version 0.9.2).
For the Neal SA sampler, we employed the \texttt{neal.SimulatedAnnealingSampler} class from the \texttt{dwave-neal} package (version 0.6.0).
For the D-Wave QA clique sampler, we utilized the \texttt{dwave.system.samplers.DWaveCliqueSampler} class from the \texttt{dwave-sampler} package (version 1.2.0) with D-Wave's Advantage system 6.4 as its solver.
For all samplers, the \texttt{num\_reads} parameter was set to 512, and for both OpenJij SQA and Neal SA samplers, the default \texttt{num\_sweeps} value of 1000 was employed.
All experiments were conducted on a MacBook Pro 2023 equipped with Apple M2 Max processor and 96 GB of memory.

\section*{Results}

\subsection*{Prioritized removal of high-risk mislabeled instances}

\begin{figure}
\centering
\includegraphics[width=\linewidth]{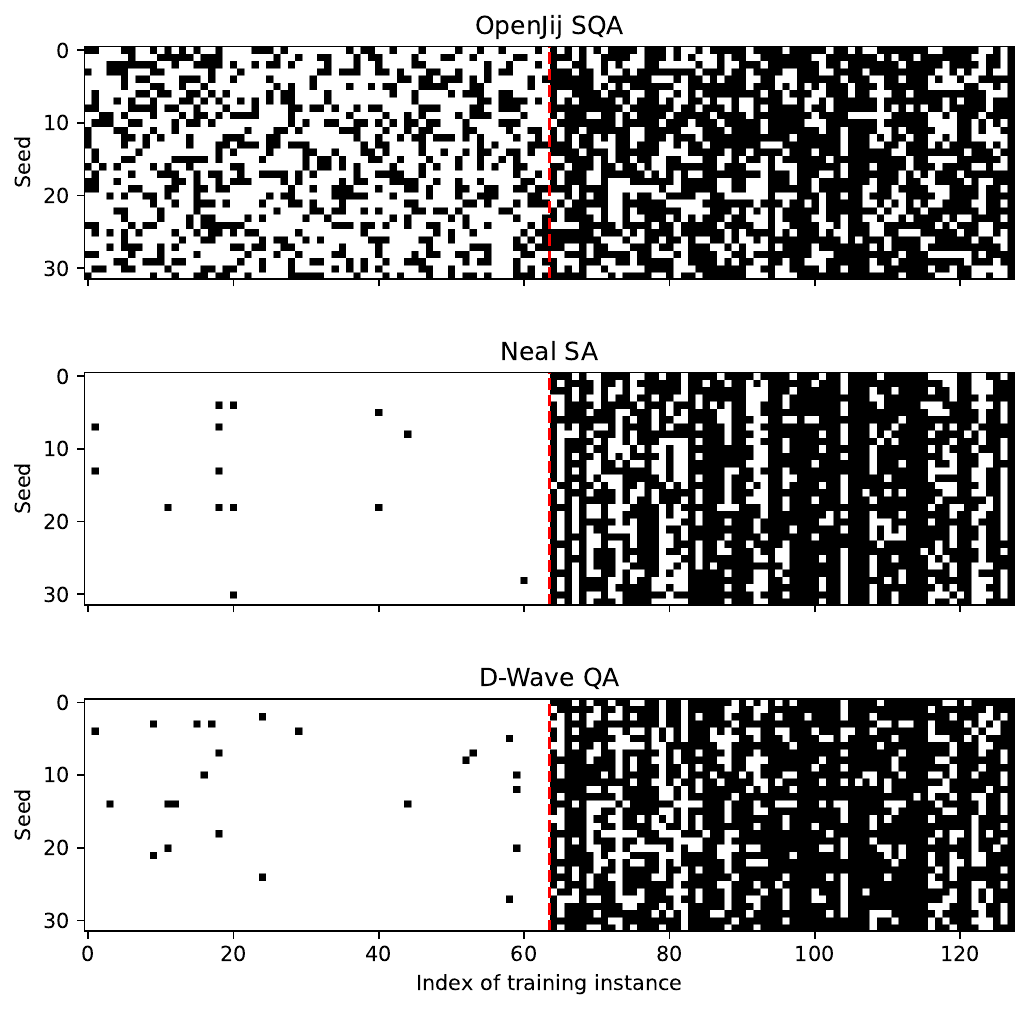}
\caption{
Instance removal patterns acquired by three different samplers over 32 runs.
Each cell in the grid represents the inclusion or exclusion status of a training instance for a specific run.
For any sampler, a white cell at the $r$-th row and $i$-th column indicates the $i$-th training instance is included in the filtered training set and used for actual training in the $r$-th run.
The black cell, on the other hand, shows the same training instance is excluded from the filtered training set and not used for training in the associated run.
The red dotted line marks the border between clean and mislabeled training instances.
Seed values are changed from 0 to 31.
This consistent filtering pattern provides clear evidence that the proposed algorithm can reliably identify mislabeled instances while preserving clean instances.
}
\label{fig:x_bests_for_different_seeds}
\end{figure}

As illustrated in Fig.~\ref{fig:x_bests_for_different_seeds}, our proposed algorithm coupled with any of three samplers consistently removed mislabeled instances (located to the right of the dotted red line in the figure) from the contaminated original training dataset, while preserving the correctly labeled ones (shown to the left of the dotted red line in the figure).
For any sampler, the subset of removed instances differed with each run, but their selection patterns remained consistent across runs.
The D-Wave QA clique sampler and the Neal SA sampler consistently kept correctly labeled instances in higher probability than the OpenJij SQA sampler.
However, regardless of this sampler-dependent performance difference in noise removal, the proposed algorithm appears to work well in removing mislabeled instances from the contaminated training dataset.

\begin{figure}
\centering
\includegraphics[width=\linewidth]{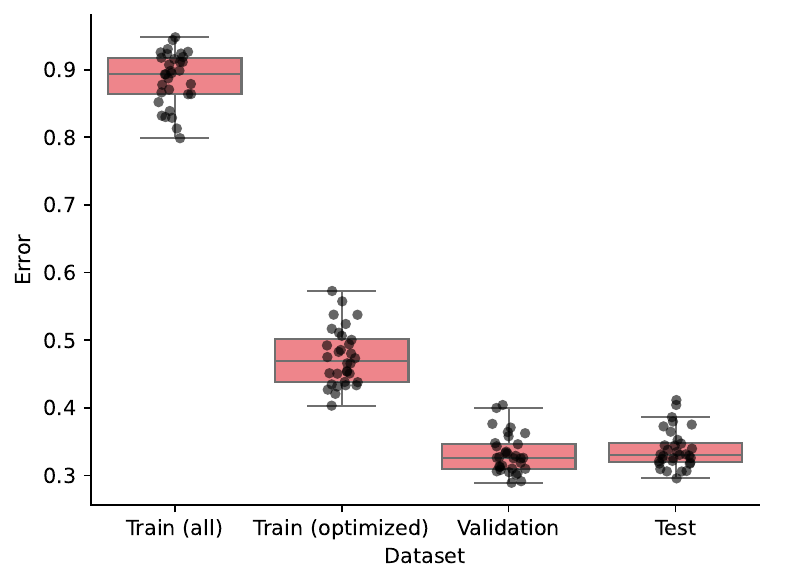}
\caption{
Performance of the trained model on four different datasets.
In each run, a model was trained using a filtered training dataset and then evaluated on four different datasets.
The model performance was measured in log-loss error metric.
Each gray dot represents one of the 32 log-loss errors for a specific dataset.
Variations in error values within the same dataset are due to differences in the training instances selected across different runs by our dataset optimization algorithm.
The labels \textit{Train (all)} and \textit{Train (optimized)} indicate whether all or filtered instances in the training dataset were used for model evaluation, respectively.
The whiskers extend to the farthest points within 1.5 times the inter-quartile range from the nearest hinge of the box plot.
The two box plots on the left confirm that training errors are consistently reduced after optimizing the training dataset in all 32 runs.
The two box plots on the right further indicate that test errors remain low and comparable to directly optimized validation errors, offering strong evidence that training with the optimized dataset enhances generalization performance.
}
\label{fig:model_performances}
\end{figure}

A logistic regression model trained on the cleansed training dataset consistently demonstrated improved performance across all 32 runs as shown in Fig.~\ref{fig:model_performances}.
Here, we focus on analyzing the results obtained using the D-Wave QA clique sampler, as similar conclusions are likely applicable to the other samplers.
The training errors computed on the optimized training subsets were consistently lower than the training errors computed on the original training sets without noise removal.
This indicates that the mislabeled instances were successfully filtered out from the contaminated training dataset.
Regardless of successful removal of the mislabeled instances, the training errors computed on the filtered training subsets remained consistently higher than the validation and test errors.
This discrepancy arises because some mislabeled instances persist in the filtered dataset even after noise removal by the proposed algorithm.
Please note that the validation error is lowest because it is used as the objective function of the surrogate model-based BBO, and all instances in the validation dataset are noise free.
The low test errors suggest that the proposed algorithm effectively removed mislabeled instances in a way to guarantee the generalization of the trained model.

\begin{figure}
\centering
\includegraphics[width=0.90\linewidth]{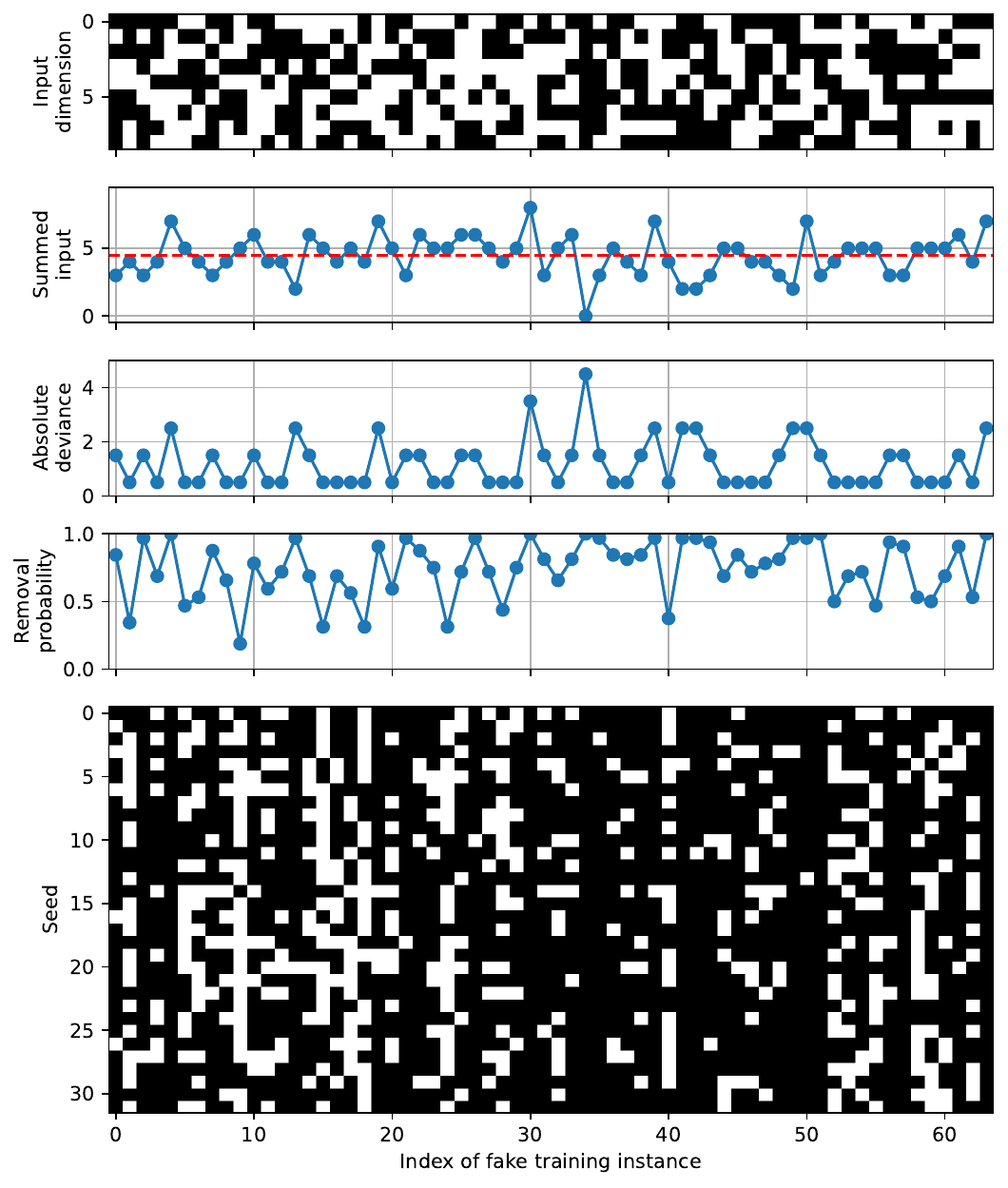}
\caption{
Characteristics of the input patterns of the fake training instances and their removal probabilities using the D-Wave QA clique sampler.
1) The 9 dimensional binary input patterns of the fake training instances.
2) The summed values of input patterns, with the horizontal dashed red line marking the optimal decision boundary of 4.5.
3) The absolute deviance (or distance) of summed input patterns from the optimal decision boundary of 4.5.
4) The removal probabilities of incorrectly labeled fake instances over 32 runs with different seed values ranging from 0 to 31.
5) The removal patterns of fake training instances across 32 different runs. This removal pattern is also depicted on the right side of the dashed red line in the D-Wave QA clique result shown in Fig.~\ref{fig:x_bests_for_different_seeds}.
The relationship between characteristics of input patterns and their removal probabilities is further analyzed in Fig.~\ref{fig:removal_probabilities_and_input_characteristics}, highlighting how input features influences the likelihood of removal.
}
\label{fig:fake_data}
\end{figure}

\begin{figure}
    \centering
    \subfigure[Summed values of input patterns]{
        \includegraphics[width=0.48\textwidth]{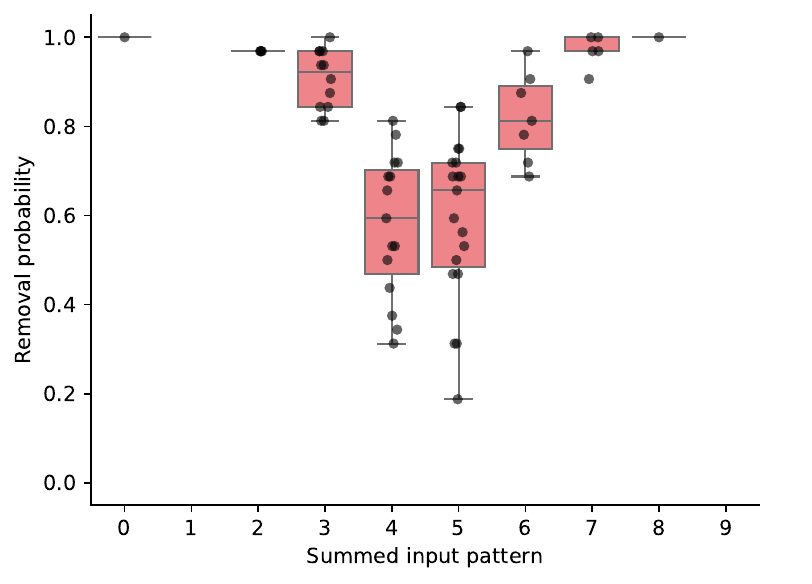}
        \label{fig:removal_probability_vs_summed_input_pattern}
    }
    \hfill
    \subfigure[Absolute deviance from the optimal threshold]{
        \includegraphics[width=0.48\textwidth]{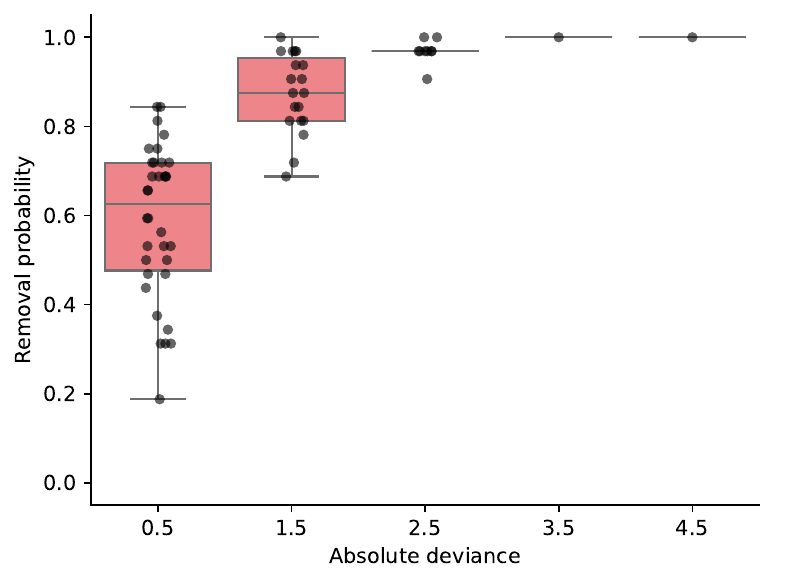}
        \label{fig:absolute_deviance_and_removal_probabilities}
    }
    \caption{
    Relationship between the removal probabilities and the characteristics of 64 input patterns of the fake training subset.
    The removal probabilities are compared with (a) the number of ones in each input pattern of the fake training subset and (b) the absolute deviance from the optimal threshold of 4.5 to the summed values of input patterns.
    In both plots, the whiskers are drawn to the farthest points within 1.5 times the inter-quartile range from the nearest hinge of the box plot.
    Both figures clearly illustrate that mislabeled training instances with greater detrimental impact on the trained model---for example, a zero vector mislabeled as class 1---are more likely to be removed.
    } \label{fig:removal_probabilities_and_input_characteristics}
\end{figure}

Further analysis revealed that mislabeled instances with a greater negative impact if not removed are more likely to be eliminated in an input-feature-dependent fashion.
Fig.~\ref{fig:x_bests_for_different_seeds} shows that the difference in removal probabilities of fake instances is persistent over different runs and different samplers.
In other words, there is a pattern in removal probability of fake instances.
Fig.~\ref{fig:fake_data} shows the 64 input vectors contained in the fake training instances and their characteristics, such as the summed value of each input vector, together with their removal probabilities computed after 32 runs with the D-Wave QA clique sampler.
The relationship between the summed values of input patterns and the removal probabilities of their associated instances is shown in Fig.~\ref{fig:removal_probability_vs_summed_input_pattern}.
Specifically, the summed input pattern, $s_{i} = \sum_{c=1}^{9} x_{i, c}$, or equivalently the probability of an element of the input vector taking the value of 1, $p_{i} = s_{i} / 9$, is shown to influence the removal probability of the input pattern $\bm{x}_{i} \in \{0, 1\}^{9}$.
To clarify this relationship, Fig~\ref{fig:absolute_deviance_and_removal_probabilities} presents the absolute deviance of summed input pattern from the optimal threshold of 4.5, which is optimal for solving the majority bit task.
This absolute deviance defined as $d(\bm{x}_{i}) = | 9 p_{i} - 4.5 |$ is computed for 64 unique input patterns in the training set $\{\bm{x}_{i}\}_{i=1}^{64}$.
Notably, this absolute deviance is inversely related to the entropy of a binary input pattern, defined as $s(\bm{x}_{i}) = - p_{i} \log p_{i} - (1 - p_{i}) \log (1 - p_{i})$.
Therefore, Fig~\ref{fig:absolute_deviance_and_removal_probabilities} demonstrates that mislabeled input patterns with lower entropy, which correspond to higher absolute deviance, are more likely to be removed.
This trend may be attributed to the stronger negative impact of low-entropy patterns with incorrect labels (e.g., $[0, 0, 0, 0, 0, 0, 0, 0, 0]$ labeled as $1$) on the trained model, compared to high-entropy patterns with incorrect labels (e.g., $[0, 1, 0, 1, 0, 1, 0, 1, 0]$ labeled as $1$).

\subsection*{Iterative improvement of solution quality}

\begin{figure}
    \centering
    \subfigure[OpenJij SQA]{
        \includegraphics[width=0.3\textwidth]{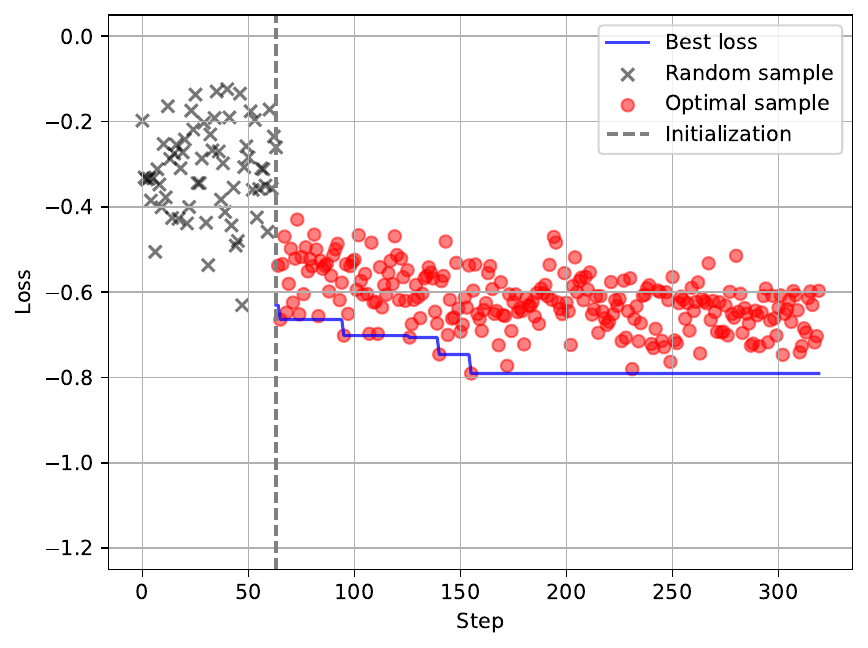}
        \label{fig:y_best_data_and_y_data_openjij_sqa}
    }
    \hfill
    \subfigure[Neal SA]{
        \includegraphics[width=0.3\textwidth]{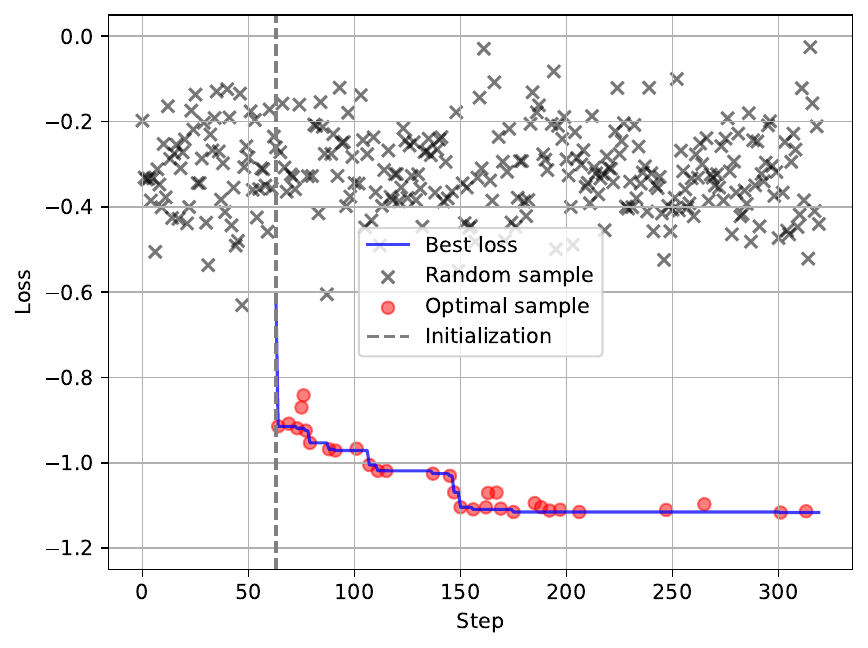}
        \label{fig:y_best_data_and_y_data_neal_sa}
    }
    \hfill
    \subfigure[D-Wave QA clique]{
        \includegraphics[width=0.3\textwidth]{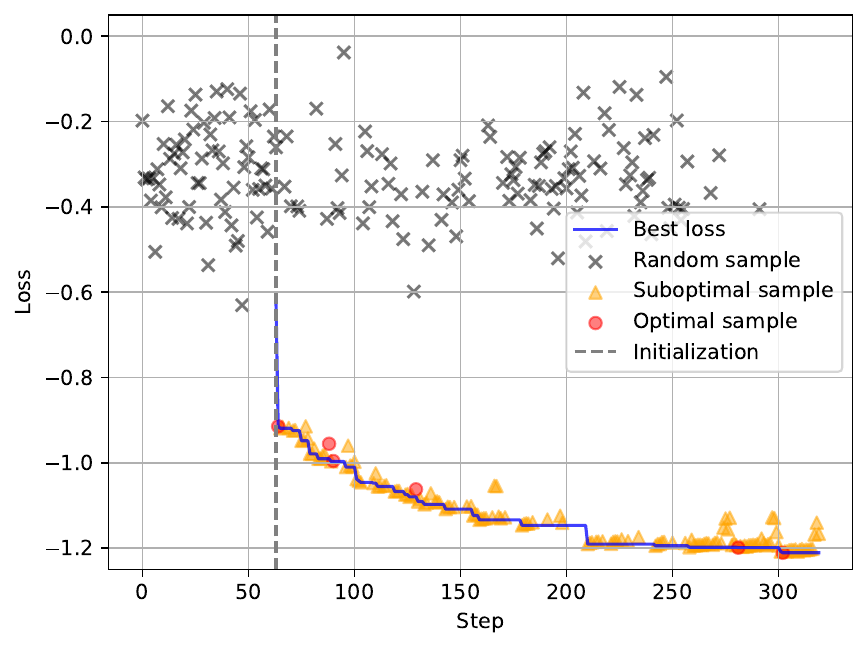}
        \label{fig:y_best_data_and_y_data_dwave_clique}
    }
    \caption{
    Comparison of the three samplers in terms of surrogate losses and the types of sequentially obtained samples.
    Each surrogate loss models the log-transformed validation loss, which takes continuous values in the real domain.
    The first 64 steps up to the dotted gray line are called the initialization phase, during which samples are obtained randomly.
    The remaining 256 steps constitute the optimization phase, where samples are obtained using the surrogate model-based BBO with postprocessing.
    In each optimization step, 512 samples are drawn from a selected sampler.
    These samples are sequentially checked from the one with the lowest energy to the one with the highest energy to see if they are already contained in the accepted sample set.
    The first sample not contained in the accepted sample set is accepted and inserted into the accepted sample set.
    If all 512 samples are already contained in the accepted sample set, a new sample not contained in the acquired sample set is drawn randomly.
    The trend of the lowest energy up to that step is shown with a blue solid line.
    A sample type is marked as \textit{optimal} if the accepted sample has the lowest energy among the 512 samples.
    A sample type is marked as \textit{suboptimal} if it is from the 512 samples but does not have the lowest energy.
    A sample type is marked as \textit{random} if it is not obtained from the 512 samples and is drawn randomly.  
    In all three cases, all parameters are fixed except for the solver types.
    As shown in Fig~\ref{fig:y_best_data_and_y_data_dwave_clique}, the D-Wave QA clique sampler effectively exploits diverse samples obtained from a physical quantum annealer, thereby achieving superior solution quality compared with classical samplers such as OpenJij SQA (Fig~\ref{fig:y_best_data_and_y_data_openjij_sqa}) and Neal SA (Fig~\ref{fig:y_best_data_and_y_data_neal_sa}) samplers.
    }
    \label{fig:y_best_data_and_y_data}
\end{figure}

\begin{figure}
    \centering
    \subfigure[Loss values of the found solutions.]{
        \includegraphics[width=0.48\textwidth]{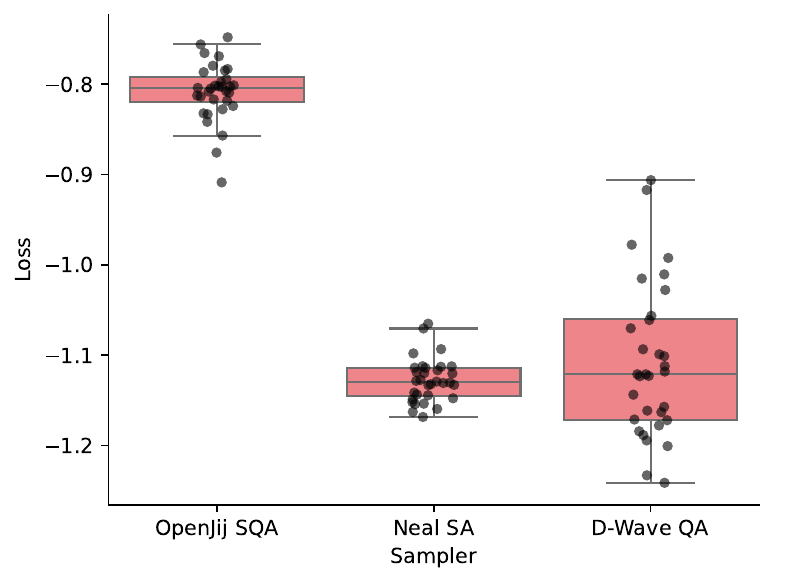}
        \label{fig:best_loss_of_different_samplers}
    }
    \hfill
    \subfigure[Hamming distances between the theoretical solution and the found solutions.]{
        \includegraphics[width=0.48\textwidth]{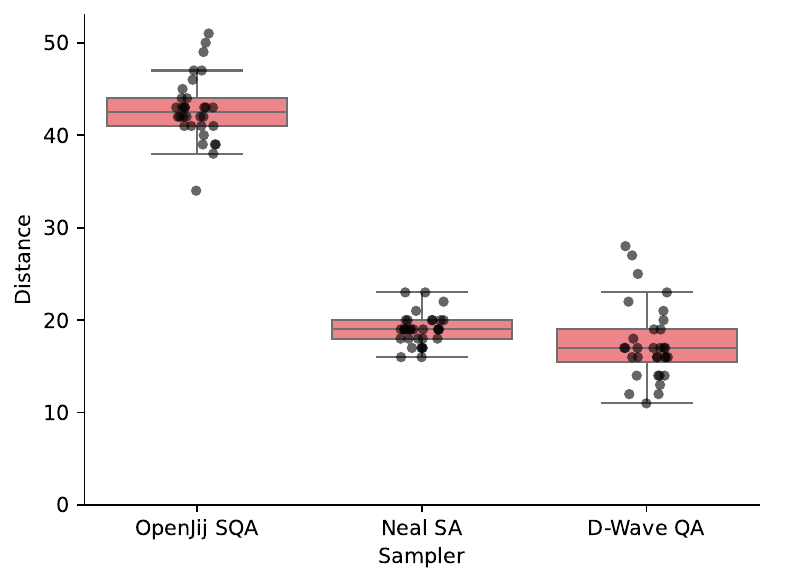}
        \label{fig:hamming_distance_for_different_samplers}
    }
    \caption{
    Characteristics of the solutions found by each sampler in 32 runs. (a) The surrogate loss value, which is the log-transformed validation loss, of the best solution found by each sampler in each run is represented as a single point. Its distribution over 32 runs is shown as a box plot for each sampler. The whiskers extend to the farthest points within 1.5 times the inter-quartile range from the nearest hinge of the box plot. (b) Hamming distance between the theoretical solution and the best solution found by each sampler in each run is shown in the same format. This Hamming distance is equivalent to the prediction accuracy of the BBO model when multiplied by the size of the training dataset.
    After collecting 96 solutions obtained by 3 different samplers over 32 runs, the D-Wave QA sampler predominantly produced the top-performing solutions in terms of both log-loss error and Hamming distance.
    } \label{fig:solution_qualities_of_three_samplers}
\end{figure}

Now let us shift our attention to compare the difference between three samplers: OpenJij SQA sampler, Neal SA sampler, and D-Wave QA clique sampler.
As shown in Fig.~\ref{fig:y_best_data_and_y_data}, in this particular run, the D-Wave QA clique sampler attained the lowest loss value followed by Neal SA and the OpenJij SQA samplers.
This tendency is not always true in different runs as shown in
Fig.~\ref{fig:solution_qualities_of_three_samplers}, but D-Wave QA clique sampler successfully found multiple solutions better than the best solution found in 32 runs of the Neal SA sampler in terms of surrogate loss (Fig.~\ref{fig:best_loss_of_different_samplers}) and Hamming distance between the theoretical solution and the found solutions (Fig.~\ref{fig:hamming_distance_for_different_samplers}).

\begin{figure}
    \centering
    \subfigure[OpenJij SQA]{
        \includegraphics[width=0.3\textwidth]{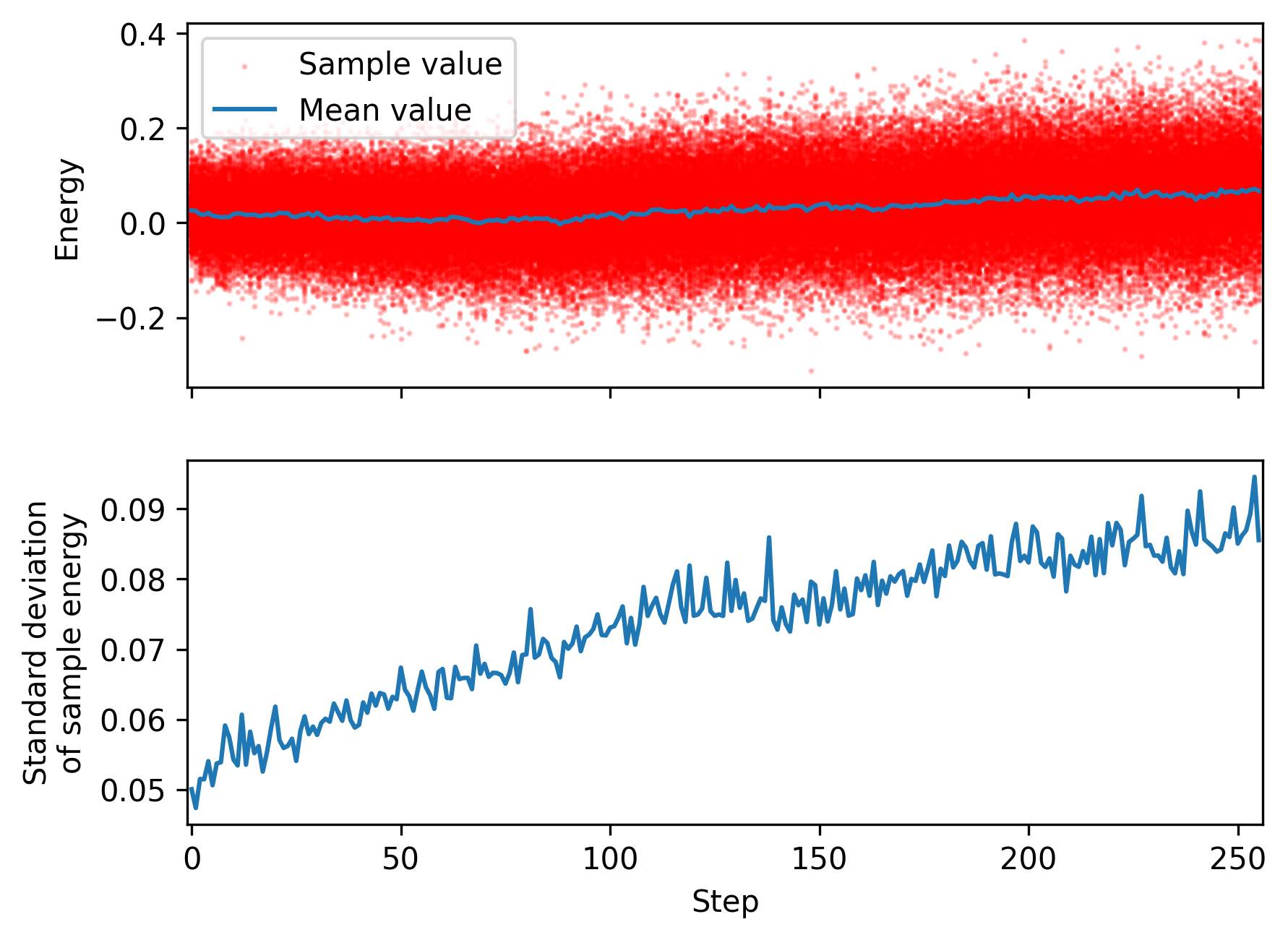}
        \label{fig:energy_trend_openjij}
    }
    \hfill
    \subfigure[Neal SA]{
        \includegraphics[width=0.3\textwidth]{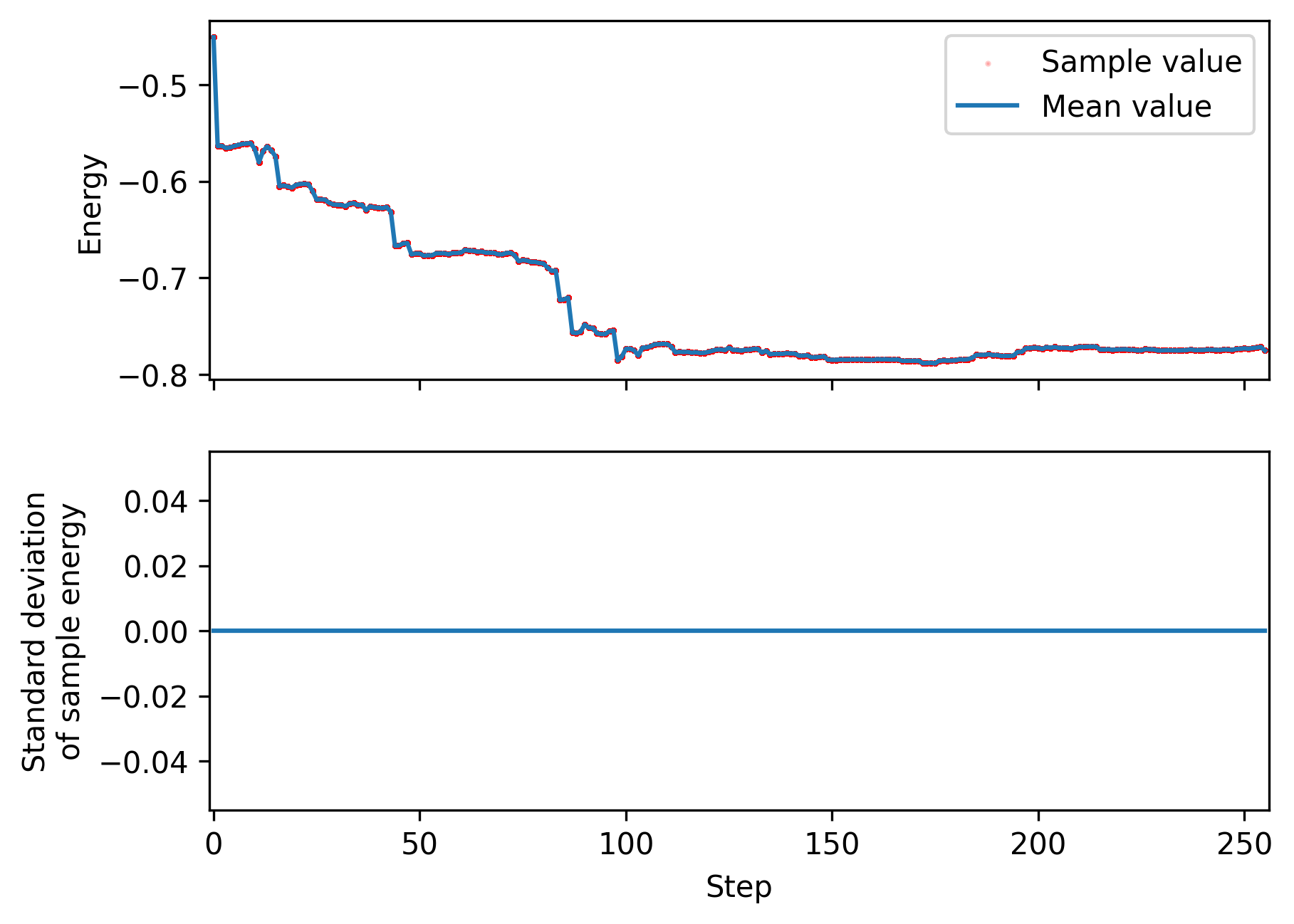}
        \label{fig:energy_trend_neal}
    }
    \hfill
    \subfigure[D-Wave QA clique]{
        \includegraphics[width=0.3\textwidth]{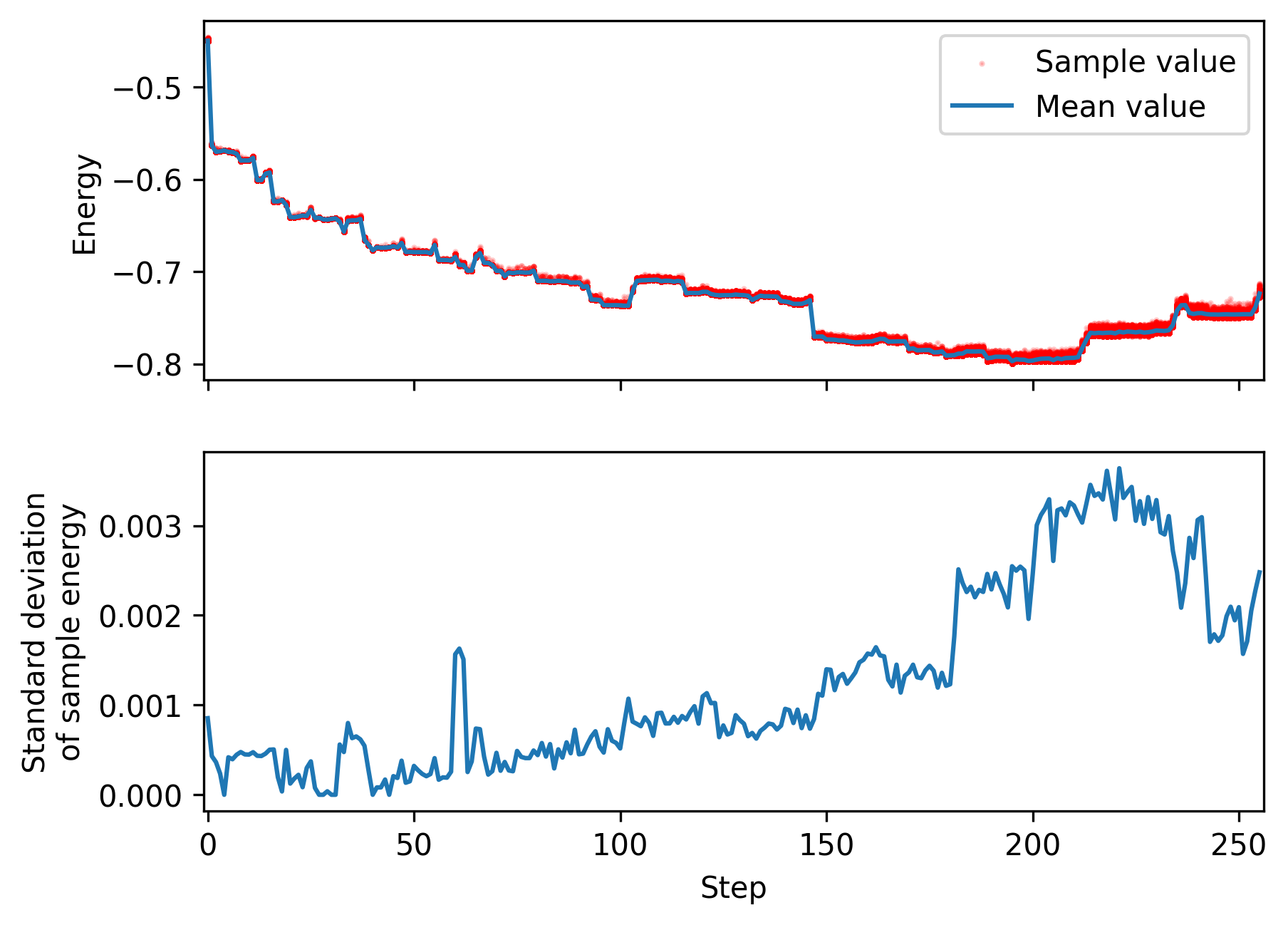}
        \label{fig:energy_trend_dwave}
    }
    \caption{
    Energy distribution of 512 samples obtained in each optimization step of three different samplers: (a) the OpenJij SQA sampler, (b) the Neal SA sampler, and (c) the D-Wave QA clique sampler.
    The top figure for each sampler shows the raw energy values of 512 samples obtained in each step over the entire 256 steps in the optimization phase.
    The mean energy value of 512 samples acquired in each step is shown with the blue line. 
    The bottom figure for each sampler shows the standard deviation of the energy values of 512 samples acquired in each step.
    Please note the energy scales of the Neal SA sampler and the D-Wave QA clique sampler are the same but different from the one used for the OpenJij SQA sampler.
    As illustrated in these figures, the three samplers have distinct sampling behaviors: (a) the OpenJij SQA sampler produces highly variable samples with elevated energy, (b) the Neal SA sampler generates low-energy but identical samples, (c) the D-Wave QA clique sampler maintains a moderate level of sample variability even in the low-energy regime.
    This distinctive characteristic of the D-Wave's physical sampler suggests that it may promote exploration in the vicinity of the global minimum.
    }
    \label{fig:energy_distributions_of_three_samplers}
\end{figure}

Fig.~\ref{fig:y_best_data_and_y_data} also shows the interesting difference in how 512 samples are utilized by each sampler in each optimization step.
Firstly, in case of OpenJij SQA sampler, the optimal sample with lowest energy in 512 samples is always accepted in each optimization step.
This is due to the fact that both the variance and the mean of energy distribution of samples obtained by OpenJij SQA sampler is always higher than ones obtained by Neal SA sampler or D-Wave QA clique sampler as shown in 
Fig.~\ref{fig:energy_distributions_of_three_samplers}.
Secondly, in case of Neal SA sampler, most of 512 samples obtained in each step are the same and have equal energy as shown in Fig.~\ref{fig:energy_trend_neal} (In this particular run, they are all equal). Therefore, there is a higher chance for the optimal sample in a certain step to be already seen before, and it causes a higher chance of random samples to be selected (See the higher probability of random samples marked by black cross shown in Fig.~\ref{fig:y_best_data_and_y_data_neal_sa}.
Thirdly, in the case of D-Wave QA clique sampler, suboptimal sample with sufficiently low energy is fully utilized to find lower energy solution in each optimization step as shown in Fig.~\ref{fig:y_best_data_and_y_data_dwave_clique}.

\subsection*{Leveraging D-Wave's quantum annealer to reach better solutions faster}

As shown in Fig.~\ref{fig:energy_trend_dwave}, 512 samples obtained by D-Wave QA clique sampler in each step have higher variety compared to ones obtained by Neal SA sampler shown in Fig.~\ref{fig:energy_trend_neal}.
This variety in 512 samples increases as the optimization phase proceeds as shown in Fig.~\ref{fig:energy_trend_dwave}.
Notice toward the end of optimization phase, the number of accepted random samples is gradually reduced as shown in Fig.~\ref{fig:y_best_data_and_y_data_dwave_clique}.

\begin{figure}
    \centering
    \subfigure[Sampling times over steps]{
        \includegraphics[width=0.48\textwidth]{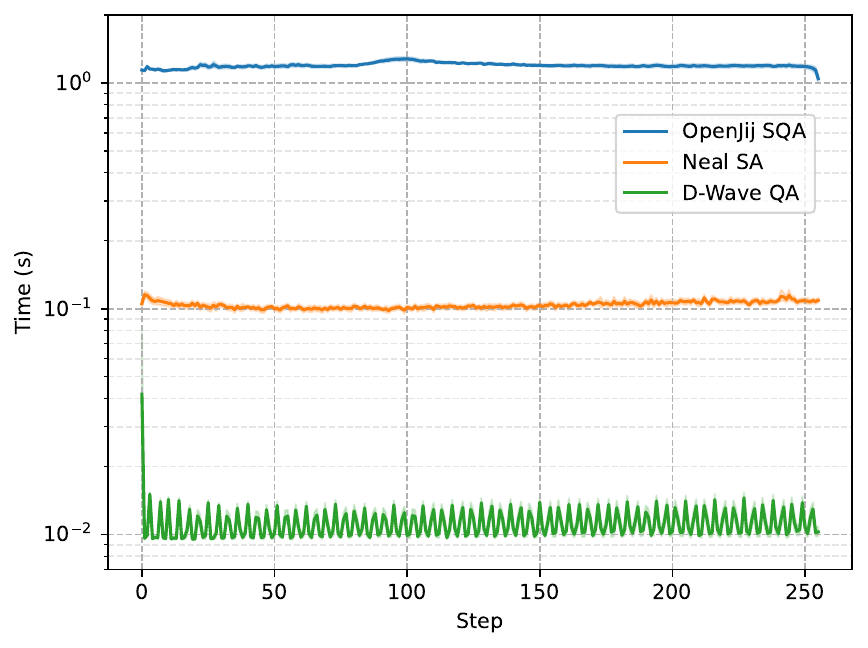}
        \label{fig:sampling_times_over_steps}
    }
    \hfill
    \subfigure[Mean sampling times]{
        \includegraphics[width=0.48\textwidth]{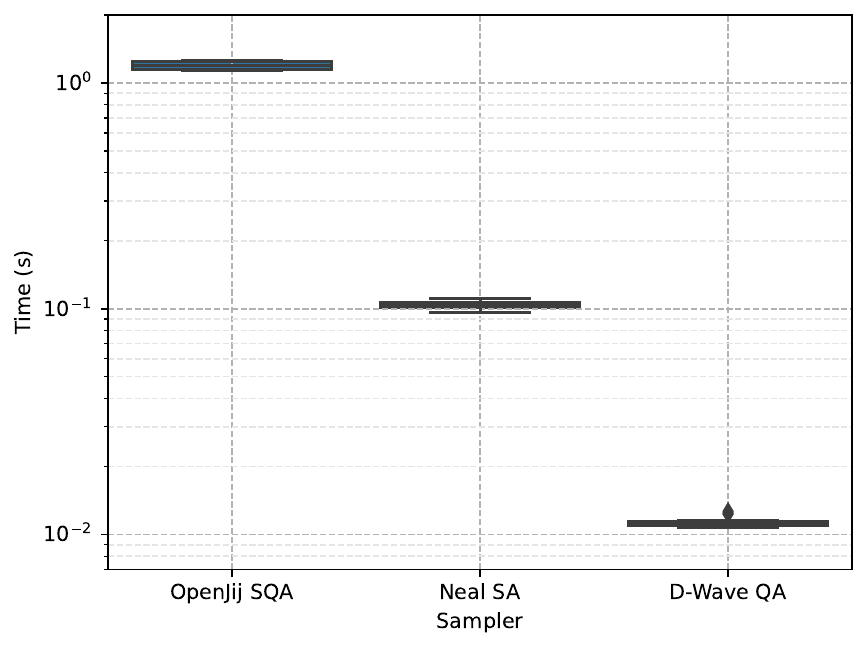}
        \label{fig:mean_sampling_times_of_three_samplers}
    }
    \caption{
    Sampling time statistics of three samplers.
    (a) Sampling times in each step are compared for the three samplers. The reported time is measured in wall-clock time and includes the time required for embedding. The 95\% confidence intervals obtained after running 32 runs are shown with transparent bands around the mean line plots.
    (b) The standard error of the mean over 32 runs, where the mean is taken over 256 steps of sampling times of each sampler.
    The whiskers extend to the farthest points within 1.5 times the inter-quartile range from the nearest hinge of the box plot.
    Overall, the mean sampling times of the OpenJij SQA sampler, the Neal SA sampler, and the D-Wave QA clique sampler are 1.196, 0.104, and 0.011 seconds, respectively.
    These results demonstrate that the D-Wave QA clique sampler achieves sampling times more than an order of magnitude faster than the other two classical samplers, highlighting its potential to obtain higher-quality solutions in substantially less time.
    }
    \label{fig:sampling_time_statistics}
\end{figure}

In addition to achieving high solution quality, the D-Wave QA clique sampler demonstrated significantly faster sampling time compared to other samplers when using the default parameter settings.
Sampling time of D-Wave QA clique sampler is about 10 times faster than that of the Neal SA sampler and about 100 times faster than that of the OpenJij SQA sampler, as shown in Fig.~\ref{fig:sampling_time_statistics}.
To assess the robustness of these differences, we conducted pairwise Mann--Whitney--Wilcoxon tests across the 32 runs.
All pairwise comparisons yielded highly significant results ($p = 6.511 \times 10^{-12}$, $U=1.024 \times 10^{3}$), confirming that the observed superiority of the D-Wave QA clique sampler over classical samplers is statistically significant. 

\section*{Discussion and Conclusion}

The proposed algorithm successfully identified mislabeled instances contained in the contaminated training dataset, as demonstrated by the substantial improvements in generalization performance.
Since the noisy majority-bit task is unsolvable above chance level without removing mislabeled instances, the improved test performance clearly indicates successful noise removal.
This success was made possible by directly optimizing validation loss via surrogate model-based black-box optimization (BBO) with postprocessing and leveraging the unique sampling capabilities of quantum annealing.

One of the key insights from this study is the observed prioritization of harmful mislabeled instances.
The algorithm was more likely to remove mislabeled samples with low input entropy, which tend to exert a stronger negative effect on generalization.
This behavior naturally emerged from the use of validation loss as the optimization objective and reflects the ability of the surrogate model-based BBO to exploit structural patterns in the data that correlate with label correctness.

More critically, we found that quantum annealing offers distinct advantages in this specific noise-removal task, beyond what is achievable with classical samplers.
In high-dimensional combinatorial problems such as instance selection (where the search space contains $2^{n}$ subsets), exploration efficiency is paramount.
Classical samplers like simulated annealing (SA) and simulated quantum annealing (SQA) often converge to narrow solution basins and suffer from sample degeneracy, limiting the breadth of search.

In contrast, D-Wave’s quantum annealer consistently produced diverse and energetically favorable solutions in each optimization step (Fig.~\ref{fig:energy_trend_dwave}), resulting in lower validation loss (Fig.~\ref{fig:y_best_data_and_y_data_dwave_clique}) and better alignment with the theoretical optimum (Fig.~\ref{fig:solution_qualities_of_three_samplers}).
This is likely due to quantum tunneling effects, which enable the system to escape shallow local minima and explore disconnected regions of the solution space more effectively than classical thermal processes.

Moreover, the reduced reliance on random sampling in later stages of optimization (Fig.~\ref{fig:y_best_data_and_y_data_dwave_clique}) and the high likelihood to find the lower surrogate losses (Fig.~\ref{fig:best_loss_of_different_samplers}) highlight the practical advantage of quantum annealing in iterative optimization under sampling constraints.
These benefits were achieved with significantly lower sampling time compared to classical samplers (Fig.~\ref{fig:sampling_time_statistics}), enhancing both solution quality and computational efficiency.

Taken together, these results suggest that quantum annealing is particularly well-suited for subset selection tasks in machine learning, especially when the solution space is vast and multi-modal.
Its ability to produce diverse, high-quality candidates at low latency offers a compelling alternative to traditional methods that rely on slower or more redundant sampling strategies.

Looking ahead, the proposed algorithm can be extended in several directions. First, by introducing binary selection over features as well as instances, it can support joint instance and feature selection within the same framework.
Second, while this study focused on synthetic data for clarity, evaluating the approach on real-world datasets with naturally occurring label noise is essential for assessing practical utility.
Finally, the method may be adapted to unsupervised learning tasks by designing suitable surrogate objectives, such as approximations to negative log-likelihood.

While the proposed algorithm demonstrates promising results in a well-controlled setting, there are several limitations in this study.
First, the experimental  task--namely the noisy majority-bit task--is synthetic and relatively small in scale, consisting of only 128 instances.
This simplified setting was deliberately chosen to allow clear analysis and control over the noise structure.
However, we recognize that such a task may not fully reflect the complexity of real-world datasets, such as CIFAR-10N/100N~\cite{wei2021learning} or Clothing1M~\cite{xiao2015learning}, where label noise occurs in more diverse and less structured forms.
Evaluation on large-scale, real-world benchmarks is necessary to assess its practical utility in future studies.
Second, we adopted a logistic regression model with $L_{2}$ regularization as a base learner,  given prior knowledge about the task.
While this choice was sufficient for our controlled study, alternative models such as LASSO or deep neural networks may be more suitable in practical real-world applications.
Finally, the applicability of the quantum annealer is constrained by its hardware limitations, particularly the number of available qubits.
If the size of the training dataset exceeds the embedding capacity of the annealer, alternative sampling strategies or a hybrid approach must be considered~\cite{mcgeoch2020}.
In addition, current quantum annealers suffer from sparse connectivity between qubits, which often requires complex minor embedding procedures that introduce additional overhead and reduce effective problem size~\cite{choi2008minor}.
This embedding overhead, combined with limited connectivity, can significantly degrade solution quality and sampling efficiency.
Moreover, thermal noise and analog control errors in the annealing process can lead to deviations from the ideal ground-state solution, further limiting reliability in practical use cases~\cite{albash2018adiabatic}.
Despite these limitations, we believe the proposed study provides a useful proof-of-concept that motivates further investigation into robust sample selection in noisy-label settings, especially in combinatorial spaces where standard gradient-based approaches may be less effective.

In this study, we have proposed a novel algorithm capable of removing mislabeled instances from a given training dataset.
This algorithm estimates the loss against a correctly-labeled validation dataset using surrogate model-based BBO followed by postprocessing.
Additionally, we have demonstrated that the proposed algorithm can be further enhanced in both solution quality and sampling speed by leveraging the unique characteristics of D-Wave Systems' quantum annealer, such as its sampling diversity and rapid sampling capability.
The proposed algorithm provides a new direction for improving the quality of training datasets by combining the power of postprocessing-equipped BBO and quantum annealing.
This combination offers a promising approach to tackling the long-standing challenge of data quality improvement.
In the future, it is important to evaluate the proposed algorithm on real-world datasets to assess its effectiveness in practical applications.
This evaluation will help determine the tangible benefits and potential limitations of the algorithm when applied to complex and diverse data scenarios.

\section*{Data availability}

The datasets used and/or analyzed during the current study are available from the corresponding author on reasonable request.

\section*{Acknowledgements}

We gratefully acknowledge the support provided by the SIP/BRIDGE program. Additionally, we extend our sincere gratitude to all participants of Quantum Infinity for You: The First Session with You, held in Kumamoto (QI4U). In particular, we would like to express our deep appreciation to Dr. Tadashi Kadowaki for his invaluable guidance during the initial phase of our research and to Dr. Akira Sasao for his dedicated efforts as the local organizer of the event.

\section*{Funding}
This work was financially supported by the programs for bridging the gap between R\&D and IDeal society (Society 5.0) and Generating Economic and social value (BRIDGE) and Cross-ministerial Strategic Innovation Promotion Program (SIP) from the Cabinet Office (23836436).

\section*{Author contributions statement}

M.O.* conceptualized and designed the study, conducted numerical experiments, analyzed the results, and wrote and revised the manuscript.
K.K. assisted with conceptualization and coding, and reviewed the manuscript.
K.M. assisted with coding, wrote and revised the manuscript.
M.O. assisted with conceptualization, wrote and revised the manuscript.
All authors reviewed the results and approved the final version of the manuscript.

\section*{Additional information}

\subsection*{Competing interests}

The authors declare no competing interests.

\end{document}